\makeatletter\def\graphicscache@inhibit{true}\makeatother

\documentclass[letterpaper, 10 pt, conference]{ieeeconf}  %

\IEEEoverridecommandlockouts                              %

\usepackage[utf8]{inputenc}

\usepackage[hidelinks]{hyperref}
\usepackage[style=ieee,hyperref,natbib=true,backend=bibtex,firstinits,doi=false,%
     mincitenames=1,maxcitenames=2,maxbibnames=99,sorting=none,terseinits=false,hyperref=true]{biblatex}
\bibliography{references.bib}
\renewbibmacro*{bbx:savehash}{}%
\defbibheading{bibliography}[\bibname]{\section*{References}}

\usepackage{url}
\usepackage{graphicx}
\usepackage[usenames,dvipsnames]{xcolor}
\usepackage[draft]{fixme}
\fxsetup{theme=color}

\usepackage{amsmath}
\usepackage{amssymb}
\usepackage{textcomp}
\usepackage{siunitx}

\usepackage{booktabs}
\usepackage{threeparttable}
\usepackage{multirow}

\usepackage[capitalize]{cleveref}

\usepackage{tikz}
\usetikzlibrary{arrows}
\usetikzlibrary{positioning,calc}
\usetikzlibrary{decorations.pathreplacing}
\usetikzlibrary{decorations.markings}
\usetikzlibrary{fit}
\usetikzlibrary{shapes.arrows}
\usetikzlibrary{shapes.geometric}
\usetikzlibrary{matrix}

\usepackage{pgfplots}
\pgfplotsset{compat=1.9}
\usepgfplotslibrary{groupplots}
\usepgfplotslibrary{units}

\usepackage[export]{adjustbox}

\IfFileExists{graphicscache.sty}{\usepackage[qfactor=0.16]{graphicscache}}

\title{\LARGE \bf
Synthetic-to-Real Domain Adaptation\\using Contrastive Unpaired Translation
}

\author{Benedikt T. Imbusch$^{1}$, Max Schwarz$^{1}$, and Sven Behnke$^{1}$%
\thanks{$^{1}$All authors are with Autonomous Intelligent Systems, University of Bonn, Germany. {\tt\small benedikt.imbusch@uni-bonn.de}}%
}

\begin{document}

\maketitle
\thispagestyle{empty}
\pagestyle{empty}
\begin{tikzpicture}[remember picture,overlay]
  \node[anchor=north,align=center,font=\sffamily\small,yshift=-0.4cm] at (current page.north) {%
  \textbf{Accepted final version.} 18th IEEE International Conference on Automation Science and Engineering (CASE), Mexico City, Mexico, August 2022
  };
\end{tikzpicture}%

\begin{abstract}

  The usefulness of deep learning models in robotics is largely dependent on
  the availability of training data. Manual annotation of training data
  is often infeasible. Synthetic data is a viable alternative, but suffers from
  domain gap.
  We propose a multi-step method to obtain training data without manual annotation effort: 
  From 3D object meshes, we generate images using a modern synthesis pipeline.
  We utilize a state-of-the-art image-to-image translation
  method to adapt the synthetic images to the real domain, minimizing the domain gap in a learned manner. The translation network is trained
  from unpaired images, i.e. just requires an un-annotated collection of real images.
  The generated and refined images can then be used to train deep learning models for a particular task.
  We also propose and evaluate extensions to the translation method that further increase
  performance, such as patch-based training, which shortens training time and increases
  global consistency.
  We evaluate our method and demonstrate its effectiveness on two robotic datasets.
  We finally give insight into the learned refinement operations.

\end{abstract}

\section{Introduction}

Robotic systems need to address several key challenges in order to be able to autonomously act in complex environments. Among these are computer vision tasks like semantic segmentation, object recognition, and 6D pose estimation. Nowadays, these tasks are most commonly solved using deep learning techniques. With increasing computation resources available, more complex network architectures are developed, raising the need for increasing amounts of training data. Acquiring training data, however, often involves tedious manual annotation of images with semantic labels or 6D poses. It is typically not feasible to create custom datasets for every specific task at hand.

To overcome this issue, previous approaches successfully relied on fine-tuning of networks pre-trained on generic datasets, reducing the required annotation effort~\citep{schwarz2018fast,morrison2018cartman}. Recently, approaches were introduced that generate synthetic training images, e.g. from 3D object meshes like \textit{Stillleben}~\citep{schwarz2020stillleben}. The benefit of such techniques is that ground truth data like 6D object poses or semantic segmentation masks are trivially available from the renderer, eliminating the need for manual annotation while providing highly accurate annotations. Although Stillleben yields good generalization to real test images on the \textit{YCB-Video} dataset~\citep{xiang2018posecnn} for semantic segmentation~\citep{schwarz2020stillleben}, the achieved results are still considerably inferior compared to training on real images. The reason for this difference is the so-called domain gap between synthetic and real data, i.e. the discrepancy between the synthetic data distribution and the real data distribution. Therefore, the model learned by a segmentation network trained on synthetic data is able to only partly capture the real data distribution from which test data is sampled.

\begin{figure}
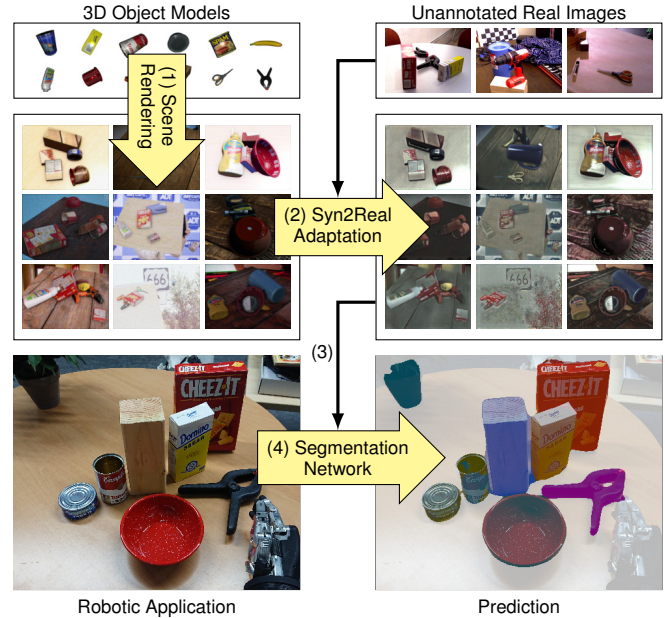

  \centering\newlength{\boxw}\setlength\boxw{3.8cm}
  \begin{tikzpicture}[
     font=\sffamily\scriptsize,
     b/.style={draw, minimum width=\boxw, minimum height=3cm,inner sep=0},
     every outer matrix/.append={b}
   ]
   \node[b,matrix,matrix of nodes,nodes={draw=none,minimum width=0, minimum height=0},row sep=2pt, column sep=2pt,minimum height=1cm] (a) {
     \includegraphics[height=.1\boxw]{figures/meshes/002_master_chef_can.jpg}       &
     \includegraphics[height=.1\boxw]{figures/meshes/004_sugar_box.jpg}             &
     \includegraphics[height=.1\boxw]{figures/meshes/005_tomato_soup_can.jpg}       &
     \includegraphics[height=.1\boxw]{figures/meshes/007_tuna_fish_can.jpg}         &
     \includegraphics[height=.1\boxw]{figures/meshes/010_potted_meat_can.jpg}       &
     \includegraphics[height=.1\boxw]{figures/meshes/011_banana.jpg}                \\
     \includegraphics[height=.1\boxw]{figures/meshes/021_bleach_cleanser.jpg}       &
     \includegraphics[height=.1\boxw]{figures/meshes/025_mug.jpg}                   &
     \includegraphics[height=.1\boxw]{figures/meshes/035_power_drill.jpg}           &
     \includegraphics[height=.1\boxw]{figures/meshes/036_wood_block.jpg}            &
     \includegraphics[height=.1\boxw]{figures/meshes/037_scissors.jpg}              &
     \includegraphics[height=.1\boxw]{figures/meshes/051_large_clamp.jpg}           \\
   };
   \node[b,right=of a,matrix,matrix of nodes,nodes={draw=none,minimum width=0, minimum height=0},row sep=2pt, column sep=2pt, minimum height=1cm] (b) {
     \includegraphics[width=.3\boxw]{figures/ycb_real/real000007.png} &
     \includegraphics[width=.3\boxw]{figures/ycb_real/real000008.png} &
     \includegraphics[width=.3\boxw]{figures/ycb_real/real000009.png} \\
   };
   \node[b,below=0.2cm of a,matrix,matrix of nodes,nodes={draw=none,minimum width=0, minimum height=0},row sep=2pt, column sep=2pt] (c) {
     \includegraphics[width=.3\boxw]{figures/syn_cut_pairs/00_rgb.png} &
     \includegraphics[width=.3\boxw]{figures/syn_cut_pairs/01_rgb.png} &
     \includegraphics[width=.3\boxw]{figures/syn_cut_pairs/02_rgb.png} \\
     \includegraphics[width=.3\boxw]{figures/syn_cut_pairs/03_rgb.png} &
     \includegraphics[width=.3\boxw]{figures/syn_cut_pairs/04_rgb.png} &
     \includegraphics[width=.3\boxw]{figures/syn_cut_pairs/05_rgb.png} \\
     \includegraphics[width=.3\boxw]{figures/syn_cut_pairs/06_rgb.png} &
     \includegraphics[width=.3\boxw]{figures/syn_cut_pairs/07_rgb.png} &
     \includegraphics[width=.3\boxw]{figures/syn_cut_pairs/08_rgb.png} \\
   };
   \node[b,right=of c,matrix,matrix of nodes,nodes={draw=none,minimum width=0, minimum height=0},row sep=2pt, column sep=2pt] (d) {
     \includegraphics[width=.3\boxw]{figures/syn_cut_pairs/00_cut.png} &
     \includegraphics[width=.3\boxw]{figures/syn_cut_pairs/01_cut.png} &
     \includegraphics[width=.3\boxw]{figures/syn_cut_pairs/02_cut.png} \\
     \includegraphics[width=.3\boxw]{figures/syn_cut_pairs/03_cut.png} &
     \includegraphics[width=.3\boxw]{figures/syn_cut_pairs/04_cut.png} &
     \includegraphics[width=.3\boxw]{figures/syn_cut_pairs/05_cut.png} \\
     \includegraphics[width=.3\boxw]{figures/syn_cut_pairs/06_cut.png} &
     \includegraphics[width=.3\boxw]{figures/syn_cut_pairs/07_cut.png} &
     \includegraphics[width=.3\boxw]{figures/syn_cut_pairs/08_cut.png} \\
   };

   \node[b,below=0.2cm of c,inner sep=0,label=south:Robotic Application] (e) {\includegraphics[width=\boxw,clip,trim=100 0 100 0]{figures/seg_img_rescaled.png}};
   \node[b,right=of e,inner sep=0,label=south:Prediction] (f) {\includegraphics[width=\boxw,clip,trim=100 0 100 0]{figures/seg_img_combined.png}};
   
   \node[anchor=south,inner sep=1pt, text depth=0pt] at (a.north) {3D Object Models};
   \node[anchor=south,inner sep=1pt, text depth=0pt] at (b.north) {Unannotated Real Images};

   \node[align=center,draw,single arrow,rotate=270,fill=yellow!50] at ($(a.south)!0.5!(c.north)$) {(1) Scene\\Rendering};
   \node[align=center,draw,single arrow,fill=yellow!50] at ($(c.east)!0.5!(d.west)$) (s2r) {(2) Syn2Real\\Adaptation};
   \node[align=center,draw,single arrow,fill=yellow!50] at ($(e.east)!0.5!(f.west)+(0,0.2)$) (segm) {(4) Segmentation\\Network};

   \draw[-latex,very thick] (b.west)  -| (s2r);
   \draw[-latex,very thick] (d.west) ++(0,-1) -| node[left,pos=0.7,inner sep=1pt] {(3)} (segm);
  \end{tikzpicture}
  \caption{
  Our method yields robust task performance in real settings, just from 3D object models
  and unannotated real images (top).
  We simulate and render plausible scenes from the 3D meshes (1).
  Our adaptation model aligns the synthetic and real image distributions more closely (2).
  The refined image dataset is used to train a task-specific network (3),
  which is applied in the target domain (4).
  None of these steps requires annotations.}
  \label{fig:teaser}
\end{figure}

We aim to obtain better results from purely synthetic data and therefore need to align the distributions more closely. We propose to perform this domain adaptation by learning a mapping from the synthetic to the real image distribution in an unsupervised manner. Specifically, this means that we only require synthetic data with ground truth and un-annotated real data without direct correspondences between the images of both datasets. To learn this mapping, we apply the GAN-based \textit{CUT} approach by \citet{park2020contrastive} in a patch-based manner. Key challenges to be addressed here include the handling of backgrounds and ensuring shape consistency. We evaluate the method on a semantic segmentation task on both the YCB-Video dataset and the \textit{HomebrewedDB} dataset~\citep{kaskman2019homebreweddb}. The individual steps of our method are visualized in \cref{fig:teaser}. To understand how the observed performance improvements can be explained, we further examine deep image features of real, synthetic, and refined synthetic frames using t-SNE embeddings.
In short, our contributions include:
\begin{enumerate}
  \item A multi-step method to obtain annotated training data from 3D object meshes and un-annotated images which can later be used for downstream applications like robotic manipulation, yielding performance close to training on real data,
  \item a patch-based application of the CUT approach to domain adaptation for two robotics datasets, and
  \item an analysis of the learned refinement operations using t-SNE embeddings.
\end{enumerate}

In the following, we denote the synthetic data distribution as $\mathcal{X}$ and a set of samples thereof as $x\in X\subsetneq\mathcal{X}$. Likewise, the real data distribution is denoted by $\mathcal{Y}$. Thus, the learned mapping can be formalized as
\begin{equation}
	f: \mathcal{X} \rightarrow \mathcal{Y}.
\end{equation}

\section{Related Work}
Domain adaptation using deep neural networks is an established area of research. \citet{wang2018deep} group the approaches into two main categories: heterogeneous and homogeneous domain adaptation. The former refers to the case when the domain gap arises from source and target domain having different feature spaces. In the latter case, both domains share their feature space but still the respective distributions $\mathcal{X}$ and $\mathcal{Y}$ do not match. In this work, we address the homogeneous case, as do the related approaches below.

\citet{stein2018genesis} apply domain adaptation to warehouse and outdoor scenes using the \textit{CycleGAN}~\citep{zhu2017cyclegan} approach. Similar to our method, they address a semantic segmentation task and
use separate networks for domain adaptation and segmentation. However, the system is applied to robotic navigation and larger-scale scene understanding, compared to robotic manipulation in small-scale scenes in our case.
This might explain why our initial experiments with CycleGAN did not yield satisfactory results.
The CUT architecture employed by our approach is easier to train and generally yields better results~\citep{park2020contrastive}.

In similar manner, \citet{mueller2018ganerated} successfully showed the use of CycleGAN-based domain adaptation for synthetic training data. Their application domain is hand pose tracking. To ensure accurate preservation of the hand poses, they propose \textit{GeoConGAN}, adding a geometric consistency loss to the CycleGAN objective. Using the newer CUT approach, we are able to avoid geometric inconsistencies by applying it in a patch-based manner. Therefore, adding complexity in the form of another loss component calculated using an additional CNN appears not justified to us.

\citet{shrivastava2017simgan} use a GAN approach with a patch-based discriminator for synthetic-to-real domain adaptation of hands and eyes with the goal of pose estimation. They use $\mathrm{L}1$-regularization on (identity or more complex) transformed image features to constrain the GAN towards content preservation. The GAN's discriminator is trained on batches of refined images accumulated over time, to stabilize the adversarial learning. CUT's contrastive learning-based approach for content consistency appears far more flexible and data-adapting to us, compared to the proposed $\mathrm{L}1$-regularization.

\citet{bousmalis2017unsupervised} propose \textit{PixelDA}, another GAN-based approach for domain adaptation. Like in our scenario, they apply it to small objects but focus more on classification and pose estimation while highlighting broader applicability. In addition to the standard setup based on a generator and a discriminator, they add a task-specific classifier to their model, trained on both synthetic and generator-refined synthetic images to support the domain adaptation. To maintain correspondences between the synthetic images and their refined versions, they propose to penalize content dissimilarities using a masked pairwise mean squared error, given depth data available from the renderer. The resulting generated backgrounds, mainly replacing black backgrounds, appear rather noisy to us. While this might even benefit generalization for classification, we expect that more consistent backgrounds are needed in our case. Additional experiments that we performed at full resolution with the CUT architecture have shown a detrimental effect of masking out the backgrounds in our application domain.

\textit{CyCADA} by \citet{hoffman2018cycada} is another domain adaptation approach derived from the idea of CycleGAN. This technique guides the adaptation process in two ways: The authors propose loss components for aligning the distributions both in the pixel space and the feature space. Besides, the authors suggest to use loss components specific to the subsequent deep learning task to enforce semantic consistency. \citet{hoffman2018cycada} report better performance on a semantic segmentation task after domain adaptation than for the existing unsupervised adaptation approaches. However, the training is computationally costly and complex---compared to the far simpler but still very effective objective of CUT.

The \textit{DLOW} technique proposed by \citet{gong2019dlow} is based on the CycleGAN concept as well. It generalizes the idea of domain adaptation beyond mapping a source domain $\mathcal{S}$ to a target domain $\mathcal{T}$: The authors introduce a model for ``domain flow generation''. Intuitively, a parameter $z\in\left[0,1\right]$ is introduced to control how far an image from $\mathcal{S}$ should be adapted towards $\mathcal{T}$. A mentioned key benefit of this technique is that learning the intermediate steps supports the domain adaptation process. In their experiments, \citet{gong2019dlow} show improved results on a semantic segmentation task compared to plain CycleGAN domain adaptation. However, the improvement is not very substantial. The previously mentioned shortcomings of CycleGAN compared to CUT apply for this approach as well.

\section{Method}
We propose a method to obtain images for the training of a deep neural network for tasks like semantic segmentation, focusing on training data for robotic manipulation.
It consists of multiple sub-steps, as illustrated in \cref{fig:teaser}: First, we generate synthetic images using the Stillleben~\citep{schwarz2020stillleben} library. Based on the image-to-image translation architecture CUT~\citep{park2020contrastive} and un-annotated real images, we then refine these synthetic images towards more realism. The refined images can then be used for training the task network. In the following, we describe the components of our approach.

\subsection{Stillleben}
The Stillleben~\citep{schwarz2020stillleben} library is
a framework for generation \& rendering of cluttered tabletop scenes.
Stillleben operates on arbitrary input meshes and generates random arrangements through the use of a physics engine.
The arranged scenes are then rendered with a modern physics-based-rendering (PBR) pipeline, producing
realistic images. A post-processing step adds effects simulating the usage of a real camera, such as
noise, chromatic aberration, white balancing errors, and over-/underexposure.
A segmentation model trained with purely Stillleben-generated synthetic data has been shown to reach
respectable performance on the YCB-Video dataset~\citep{schwarz2020stillleben}.

\subsection{Contrastive Unpaired Translation (CUT)}
Our domain adaptation approach is largely based on \textit{Contrastive Unpaired Translation (CUT)} as introduced by \citet{park2020contrastive}, which we briefly introduce here. It is an image-to-image translation technique, aimed at preserving the image content while adapting the appearance to the target domain. CUT is related to the well-known CycleGAN approach by \citet{zhu2017cyclegan} which pursues the same objective. Both are GAN-based, can be used for unpaired image sets, and have to address the same key issue: Training a GAN for unpaired image-to-image translation is in general an under-constrained task. CycleGAN employs a second GAN for a reverse mapping from the target to the source domain. The method enforces correspondences between input and output image by a cycle-consistency loss that penalizes differences resulting from passing an image through both the forward and the reverse GAN subsequently. This avoids collapse of the generator, i.e. mapping all inputs to a single
output in the target domain.

\begin{figure}
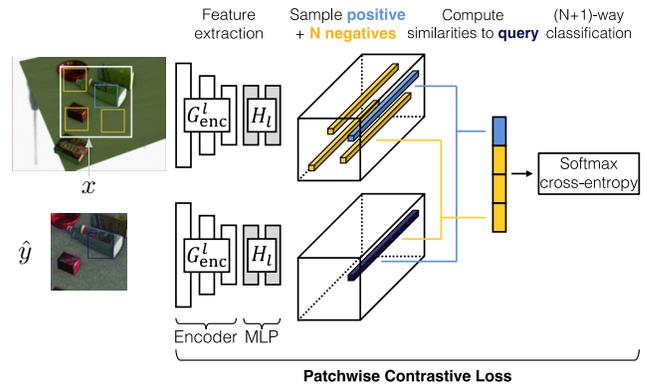
\centering
  \begin{tikzpicture}
    \node at (0,0) (x) {\includegraphics[scale=.09]{figures/cut_nce_arch_patch/input_with_patches.png}};
    \node[below=-3pt of x] (xlabel) {$x$};
    \draw[-latex,black!20!white] ($(xlabel) + (0,0.15)$) -- ($(x) + (0,-0.3)$);
    \node[below=9pt of x] (yhat) {\includegraphics[scale=.1]{figures/cut_nce_arch_patch/refined_with_subpatch.png}};
    \node[below right=-2.425cm and -4pt of x] (overview) {\includegraphics[width=.72\linewidth]{figures/cut_nce_arch_patch/cut_figure_crop.pdf}};
    \node[left=0pt of yhat] (yhatlabel) {$\hat{y}$};
  \end{tikzpicture}
  \caption{The PatchNCE is calculated based on the selected patch $x$ from the synthetic image and the corresponding generated refined image $\hat{y}$. From these smaller images, subpatches are selected for the calculation. Adapted from~\citep{park2020contrastive}.}
  \label{fig:nce_architecture_patch}
\end{figure}

CUT removes the second GAN and replaces the cycle consistency loss with a contrastive loss on image patches, the so-called \textit{PatchNCE} loss. In brief, the idea is to achieve content preservation by ensuring that a patch of the translated image has more information in common with the same patch in the source image (positive) than with $N$ other patches from the source image (negatives). Technically, this is realized by training a small MLP classifier to select the positive from the $N+1$ source patches, given the translated patch. This is done in the GAN encoder's feature space, separately for each layer used. The concept is illustrated in \cref{fig:nce_architecture_patch}. In CUT, the PatchNCE is also calculated for the image from the target distribution to stabilize the training by hinting the network to keep images from the target domain identical. The complete loss function thus is given by
\begin{align}
\begin{split}
  \mathcal{L} &= \mathcal{L}_{\mathrm{GAN}}\left(G, D, X, Y\right) \\
              &+ \lambda_{\mathrm{NCE}} \cdot (\mathcal{L}_{\mathrm{PatchNCE}}\left(G, H, X\right)
    + \mathcal{L}_{\mathrm{PatchNCE}}\left(G, H, Y\right)),  %
\end{split}
\end{align}
where $G$ denotes the GAN's generator, $D$ the discriminator, $X$ the source image and $Y$ the target image. $H$ denotes the MLPs used for the PatchNCE. \citet{park2020contrastive} suggest to choose $\lambda_{\mathrm{NCE}} = 1$.

\subsection{Modifications to CUT}
Compared to CycleGAN, we selected the CUT approach for its lesser complexity at better performance (see \citep{park2020contrastive}).

\begin{figure}
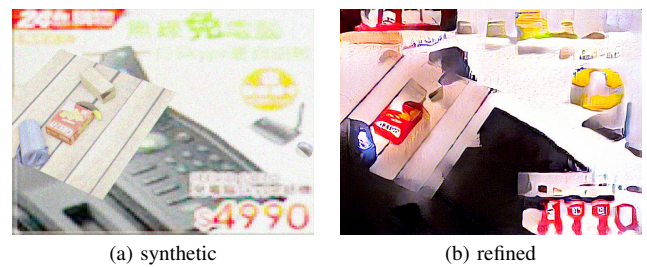

  \centering
  \begin{tikzpicture}[font=\footnotesize]
   \node[matrix,matrix of nodes,nodes={inner sep=0}, row sep=4pt, column sep=10pt] {
    \includegraphics[width=.225\textwidth]{figures/cut_fullres/sl_fullres.png} &
    \includegraphics[width=.225\textwidth]{figures/cut_fullres/cut_fullres.png} \\
    (a) synthetic & (b) refined \\
   };
  \end{tikzpicture}
  \caption{A synthetic image and a CUT-refined version of it with CUT trained on full-resolution images. Note how training at full resolution leads to deformations and hallucinations
  of objects in the background image.}
  \label{fig:cut_fullres}
\end{figure}

Its authors propose to apply CUT to images at full resolution. However, we decided to train it in a patch-based way. There are several reasons for this:
First, we have substantial variability in the images produced by Stillleben, especially with respect to the backgrounds, and a large set of real images. It seems sensible to us to use many of these images for the training in order to achieve good generalization to unseen images.
Working at full resolution (640$\times$480 for YCB-Video), however, induces long training times when following the learning duration of 400 epochs proposed by \citet{park2020contrastive}.
Second, the changes to the source image we aim to achieve are at small scale. Ideally, our domain-adapted images reflect the visual properties induced by the camera used for the real images but are content-wise very close to the source data to keep the segmentation labels usable.
Experimental results support this motivation. CUT models trained at full resolution often deform relevant objects or hallucinate parts of them in new places, as can be seen in \cref{fig:cut_fullres}. While the training is performed on patches, inference is still possible at full resolution, thanks to the GAN generator's architecture.
We thus argue that it is sufficient and even beneficial to work on image patches.

The selected patch size has to be small enough to notably reduce the computation effort and prevent global effects.
At the same time, it has to be large enough to still contain sufficient information and to ensure that sampling sub-patches for the PatchNCE is still possible in a meaningful way. We propose and evaluate patch sizes between $60^{2}$ and $160^{2}$ pixels. The patch selection is done by random cropping.

\section{Evaluation}
\subsection{Experimental Setup and Evaluation Metric}
In many applications, the goal of visual domain adaptation is to create images that look appealing to the human eye. In our robotics use case, however, we are not mainly interested in well-looking images but in images that yield better results on subsequent data processing tasks than the original synthetic images.

Therefore, we evaluate our method on a semantic segmentation task similar to how it is done in~\citep{schwarz2020stillleben}: We use \textit{RefineNet}~\citep{lin2017refinenet}, an established network architecture for semantic segmentation, and train it from scratch on 450k images, subdivided into 300 epochs of 1500 images. The segmentation performance is evaluated on a test set of annotated real images by calculating the mean intersection over union (mean IoU or mIoU) over all classes present in the dataset. All IoU values in the following are calculated on the respective test sets.

We are interested in the performance on test data after training on three image sets: synthetic images from Stillleben, CUT-refined synthetic images from Stillleben, and also real images (disjoint from the test set) for comparison. For the refined images, we use the ground-truth labels provided by the renderer for the corresponding unrefined images. Ground-truth labels for both the training and test real images are provided as part of the datasets used for the evaluation.

\begin{figure}
  \centering
  \begin{tikzpicture}
    \begin{axis}[
      xmin=0, xmax=300, width=\linewidth, height=6cm,
      y label style={rotate=0},
      xlabel={training epoch},
      ylabel={IoU}
    ]
      \addplot[thin,blue] table {figures/train_iou-syn.txt};
    \end{axis}
  \end{tikzpicture}
  \caption{Test IoU on YCB-Video for training on synthetic images over 300 epochs shows significant variance between epochs.}
  \label{fig:iou_differences}
\end{figure}
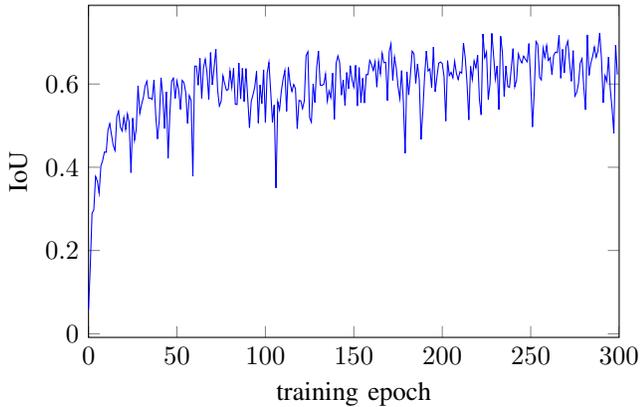

As can be seen in \cref{fig:iou_differences}, the performance on the test data differs significantly between epochs during the training of RefineNet. In the following, we therefore always visualize the distribution of IoU values over the $50$ last training epochs instead of just indicating the IoU value for the final epoch.

To obtain the refined images, we pass the synthetic images through the CUT generator. Before, we train CUT on unpaired sets of synthetic and real images for 400 epochs, following the curriculum suggested by \citet{park2020contrastive}. For the training of CUT, we choose a batch size of $40$---irrespective of the patch size. This is to keep the sources of variation between the results for different patch sizes limited. The deviation from the standard batch size of $1$ for CUT has two main reasons: First, the training time can be substantially reduced while improving memory usage. Second, we argue that for the patch-based application of CUT it is beneficial to use more averaged gradients, especially as some of the randomly sampled patches may consist entirely of irrelevant background information. Besides, we do not horizontally flip images for data augmentation to only expose the network to images that could occur in reality. Further deviations from the defaults of the standard implementation are stated in the respective experiment descriptions for the two datasets considered. All training of CUT has been performed using NVIDIA A100-SXM4-40GB accelerator cards.

We mainly work with the YCB-Video dataset but demonstrate the broader applicability of our approach on the HomebrewedDB dataset as well.

\subsection{Results on YCB-Video}
Stillleben has been evaluated on the YCB-Video dataset~\citep{schwarz2020stillleben}, thus this is the best-suited dataset for a comparative evaluation. For the training of CUT, we use 10k images generated using Stillleben and 10k images from the training set of YCB-Video.

\begin{figure}
  \centering
  \includegraphics[width=\linewidth]{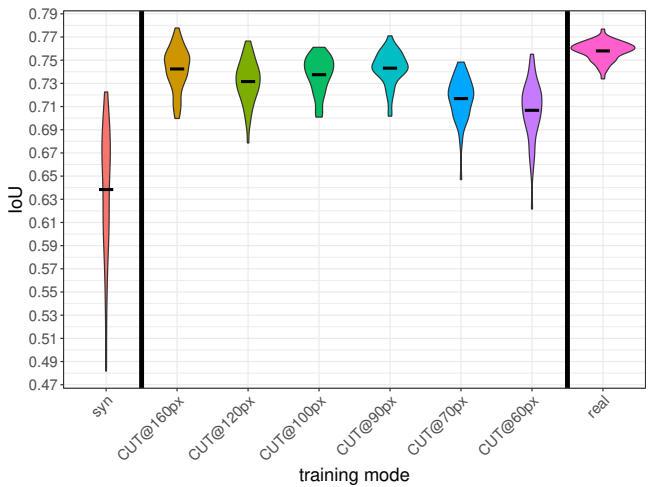}
  \caption{Results on YCB-Video. The test IoU distribution over the last $50$ training epochs for CUT-refined images for CUT patch sizes $160^{2}$, $120^{2}$, $100^{2}$, $90^{2}$, $70^{2}$, and $60^{2}$. The leftmost plot depicts the test results for training with synthetic images, the rightmost plot for training with real images. Training with CUT-refined synthetic images not only yields higher IoU values than pure synthetic images but also narrower distributions.} %
  \label{fig:iou_ycb_patchsizes}
\end{figure}

The first investigated aspect is the patch size, which we choose between $60^{2}$ and $160^{2}$ pixels as mentioned above. It is worth noting that internally, CUT works with patch sizes that are multiples of $4$. Otherwise, patches are rescaled to have such size using bicubic interpolation. While it appears possible that this interpolation would introduce some beneficial or detrimental smoothing to the input images, we saw no consistent effect on the results. We evaluate the performance for the following patch sizes: $60^{2}$, $70^{2}$, $90^{2}$, $100^{2}$, $120^{2}$, and $160^{2}$ pixels. The results are shown in~\cref{fig:iou_ycb_patchsizes}, alongside the results for RefineNet trained on purely synthetic and real YCB-Video images. It can be seen that especially at patch size $60^{2}$, but to some degree still for $70^{2}$, only insufficient information is conveyed for this dataset---compared to the larger patch sizes. $160^{2}$ and $90^{2}$ appear to yield the best results. Given the fact that the training time for CUT largely depends on the patch size (see~\cref{table:cut_training_time_patchsize_ycb}), $90^{2}$ seems to be the best trade-off. Consequently, using CUT-refined synthetic images offers a significant benefit over using pure Stillleben images as the performance is close to what training on real data yields.
\begin{table}
  \caption{Patch Size \& Training Epoch Time}
  \label{table:cut_training_time_patchsize_ycb}
  \centering
  \begin{tabular}{cr}
    \toprule
    patch size [px] & time [s] \\ \midrule
    160$\times$160 & 181 \\
    120$\times$120 & 110 \\
    100$\times$100 &  82 \\
     90$\times$90 &  66 \\
     70$\times$70 &  54 \\
     60$\times$60 &  43 \\ \bottomrule
  \end{tabular}
\end{table}

Apart from the general performance increase, CUT-refining synthetic images offers another benefit: Irrespective of the patch size, refining the images leads to a significantly narrower distribution of the IoU values, closer
to real data, which yields the narrowest distribution.
In contrast, the variance is rather high for training on synthetic images.

\subsection{Modifications}
\begin{figure}
  \centering
  \includegraphics[width=\linewidth]{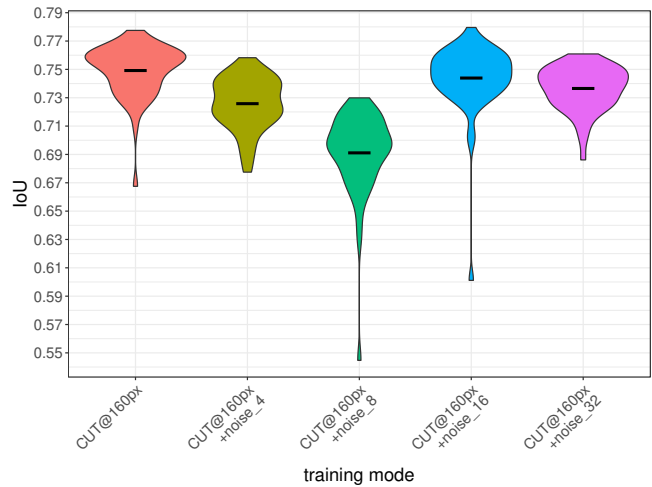}
  \caption{Noise injection. The test set IoU distributions for $0$, $4$, $8$, $16$, and $32$ injected random feature maps show no beneficial effect of injecting noise.}
  \label{fig:iou_ycb_noise}
\end{figure}

Synthetic images can imitate image noise, for instance as proposed by \citet{foi2008practical}. Noise from real cameras, however, exposes properties that are hard to model and is inherently random. It could be the case that at the synthetic-to-real domain adaptation task, it is hard for a GAN to add such noise to an image. To address this and even further increase the IoU, we tested noise injection to add noise within the translation process. Given the GAN's encoder-decoder architecture, we decided to inject it directly at the input of the decoder by adding $N$ normally-distributed random feature maps to the $M$ feature maps from the encoder, where $N \ll M$. To ease the noise integration and return to the previous number of feature maps, we add three convolution layers before the actual decoder part.
Using a patch size of $160^{2}$ pixels, we tested injecting $0$, $4$, $8$, $16$ and $32$ random feature maps. The results can be seen in~\cref{fig:iou_ycb_noise}. While $8$ feature maps appear to even have a detrimental effect across multiple runs of this experiment, the general impression is that injecting noise is not beneficial. This contradicts the hypothesis that additional randomness apart from the noise added by Stillleben's camera model helps the GAN to closer match the real image distribution. We hypothesize that the artificial image noise from Stillleben is sufficient to produce images that cannot too easily be distinguished from real images by the GAN discriminator based on the kind of noise.%

\begin{figure}
  \centering
  \includegraphics[width=\linewidth]{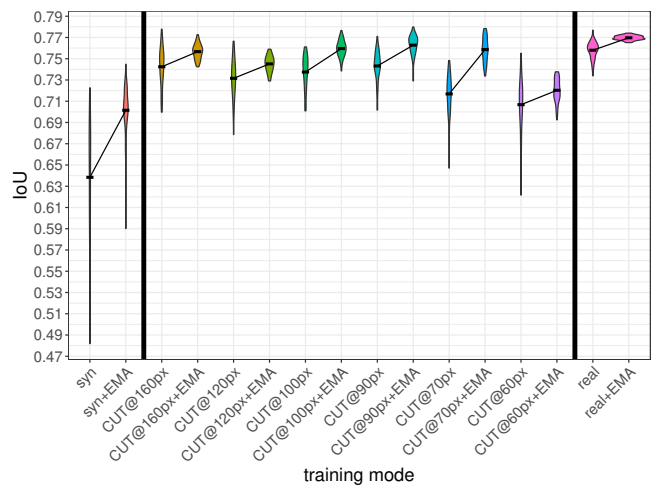}
  \caption{Using exponential moving average (EMA) for the RefineNet parameters improves the performance and reduces the IoU variability.}
  \label{fig:iou_ycb_ema}
\end{figure}

Another change however is beneficial, but not directly related to CUT: Using an exponential moving average (EMA) of the RefineNet model parameters for the evaluation on the test set does significantly improve the performance and reduce the variability of the IoU (decay factor: $0.995$). This is consistent over all patch sizes for CUT-refined images as well as synthetic and real images, as can be seen in \cref{fig:iou_ycb_ema}. Note that we achieve more than 99\% of the mean IoU for real training data, see~\cref{table:iou_ycb_ema}. We therefore use this modification in the following for the experiments with HomebrewedDB.
\begin{table}
	\caption{Results on YCB-Video (with EMA).}
	\label{table:iou_ycb_ema}
	\centering
	\begin{tabular}{lrrrr}
		\toprule
		Training mode & mIoU $\left(\uparrow\right)$ & vs. Real $\left(\uparrow\right)$ & vs. Syn. $\left(\uparrow\right)$ \\ \midrule
		synthetic~\citep{schwarz2020stillleben} +EMA & 0.701 & 0.910 & --- \\ \midrule
		$\mathrm{CUT}_{160\mathrm{px}}$ & 0.757 & 0.983 & +8.0\% \\
		$\mathrm{CUT}_{120\mathrm{px}}$ & 0.745 & 0.968 & +6.3\% \\
		$\mathrm{CUT}_{100\mathrm{px}}$ & 0.760 & 0.987 & +8.4\% \\
		$\mathrm{CUT}_{90\mathrm{px}}$  & \textbf{0.763} & \textbf{0.991} & \textbf{+8.8\%} \\
		$\mathrm{CUT}_{70\mathrm{px}}$  & 0.759 & 0.986 & +8.3\%  \\
		$\mathrm{CUT}_{60\mathrm{px}}$  & 0.720 & 0.935 & +2.7\% \\ \midrule
		real~\citep{schwarz2020stillleben} +EMA & 0.770 & 1.000 & +9.8\% \\ \bottomrule
	\end{tabular}
\end{table}

\subsection{Results on HomebrewedDB}
Encouraged by the promising results on the YCB-Video dataset, we also evaluate our approach on the HomebrewedDB~\cite{kaskman2019homebreweddb} dataset. Both datasets consist of small objects on a table. However, the overall image appearance and the presented arrangements differ by a lot. We use the HomebrewedDB data offered in the \textit{BOP} challenge\footnote{https://bop.felk.cvut.cz/datasets/}. For both the offered test images (\textit{BOP'19/20 test images (Primesense)}) and the validation images (\textit{Validation images (Primesense)}), ground truth semantic annotations are included. No real training images are provided. However, 3D meshes for all objects are available, allowing us to use Stillleben.

We restrict our evaluation to the subset $S$ of HomebrewedDB objects which are present in the test images.
We generate synthetic scenes with objects from $S$ and train CUT on validation images containing objects from $S$.
The model trained on real data is trained analogously.
In all cases, the official test set is used to evaluate the segmentation performance of the trained RefineNet models.

\begin{figure}
  \centering
  \includegraphics[width=\linewidth]{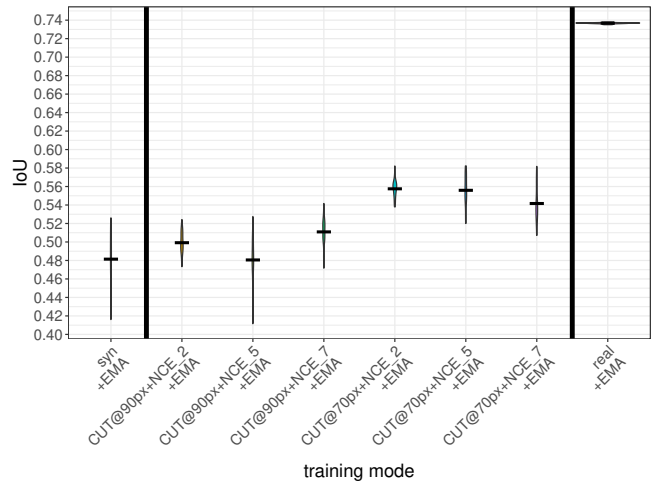}
  \caption{Results on HomebrewedDB.
  We show the test IoU distribution over the last $50$ training epochs for CUT-refined images for CUT patch size $90^{2}$ and $\lambda_{\mathrm{NCE}} = 2,\ 5,\ 7$  as well as patch size $70^{2}$ with the same values for $\lambda_{\mathrm{NCE}}$ (from left to right), complemented by the results for training on synthetic (left) and real images (right).}
  \label{fig:hbdb_iou}
\end{figure}

For generating the synthetic images using Stillleben, we make two slight modifications compared to the process proposed for YCB-Video by \citet{schwarz2020stillleben}, based on the appearance of the real images: We remove the stickers randomly added to the objects and instead of rendering the objects on a textured table, a white table is used. Without any further adaptation to HomebrewedDB, we reach IoU values around $0.5$ which are inferior to those achieved for YCB-Video. The achieved IoU for real images is slightly lower than for YCB-Video, see \cref{fig:hbdb_iou}.

For the training of CUT, we use 10k synthetic images and all 1020 real validation images. Obviously, most real images are presented multiple times during each epoch. Due to our patch-based training, the shown part of the image however varies. Based on our results on YCB-Video, we restrict ourselves to patch sizes of $70^{2}$ and $90^{2}$ pixels.

With the same hyperparameter choices as for YCB-Video, we saw segmentation performance inferior to that of pure Stillleben. Looking at the produced images, the reason is a mode collapse: Most of the images are grey-textured, with the object shapes barely visible. This is consistent over multiple runs and patch sizes.
A reason might be that for this dataset and the respective synthetic images produced by Stillleben, the loss weighting insufficiently ensures content preservation. Therefore, we propose to increase the weight of the PatchNCE while keeping it low enough to allow for meaningful changes to the appearance. The results for $\lambda_{\mathrm{NCE}} = 2,\ 5,\ 7$ can be seen in~\cref{fig:hbdb_iou} for both patch sizes considered. A modest increase of $\lambda = 2$ appears to be favorable, as does a patch size of $70^{2}$---with regards to the IoU and it's variability as well.
With these changes, we see a significant improvement over the results using raw Stillleben images, quantitatively larger than for YCB-Video (with EMA), see~\cref{table:hbdb_iou}. Not only the IoU is increased by CUT-refining the synthetic images, but also the IoU variance over the training epochs of RefineNet is reduced---consistent with what we see for YCB-Video.

\begin{table}
	\caption{Results on HomebrewedDB (with EMA).}
	\label{table:hbdb_iou}
	\centering
	\begin{tabular}{lrrrr}
		\toprule
		Training mode & Mean IoU $\left(\uparrow\right)$ & vs. Real $\left(\uparrow\right)$ & vs. Syn. $\left(\uparrow\right)$ \\ \midrule
		synthetic & 0.481 & 0.653 & --- \\ \midrule
		$\mathrm{CUT}_{90\mathrm{px},\lambda_{\mathrm{NCE}}=2}$ & 0.499 & 0.677 & +3.7\% \\
		$\mathrm{CUT}_{90\mathrm{px},\lambda_{\mathrm{NCE}}=5}$ & 0.481 & 0.653 & +0.0\% \\
		$\mathrm{CUT}_{90\mathrm{px},\lambda_{\mathrm{NCE}}=7}$ & 0.511 & 0.693 & +6.2\% \\
		$\mathrm{CUT}_{70\mathrm{px},\lambda_{\mathrm{NCE}}=2}$ & \textbf{0.558} & \textbf{0.757} & \textbf{+16.0\%} \\
		$\mathrm{CUT}_{70\mathrm{px},\lambda_{\mathrm{NCE}}=5}$ & 0.556 & 0.754 & +15.6\%  \\
		$\mathrm{CUT}_{70\mathrm{px},\lambda_{\mathrm{NCE}}=7}$ & 0.542 & 0.735 & +12.7\% \\ \midrule
		real      & 0.737 & 1.000 & +53.2\% \\ \bottomrule
	\end{tabular}
\end{table}

Thus, our approach is applicable also beyond YCB-Video with only minor changes needed for a related dataset.

\subsection{Combination with Real Training Data}
\begin{figure}
  \centering
  \includegraphics[width=\linewidth]{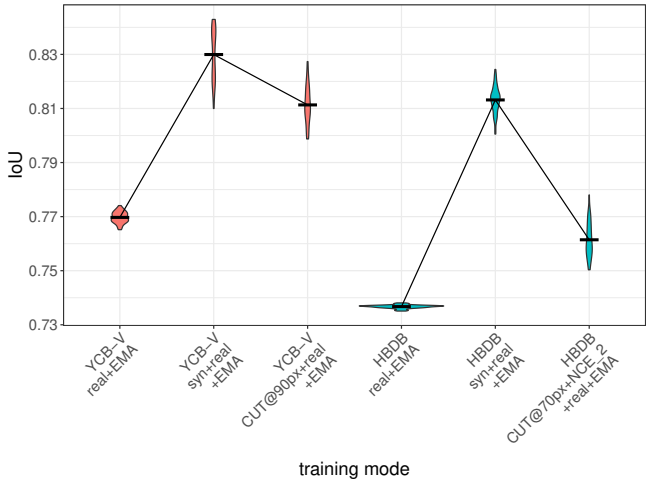}
  \caption{Mixing synthetic and real data.
  We show the test IoU distribution over the last $50$ training epochs for real training images, real and synthetic images mixed, as well as real and refined synthetic images mixed, for YCB-Video and HomebrewedDB, respectively.}
  \label{fig:iou_withreal}
\end{figure}

Until now, we considered the case where we have no real training data and aim to enhance the usefulness of synthetic data. However, there might also be cases where we have real training data available but the achieved performance is not good enough. In their experimental setup, \citet{schwarz2020stillleben} have shown that it is beneficial to train RefineNet on synthetic and real data at the same time by randomly choosing the mini-batches from both datasets. One might wonder whether the use of CUT-refined images enhances this effect. The results for both YCB-Video and HomebrewedDB are depicted in \cref{fig:iou_withreal}. While the achieved IoU on the respective test sets is still higher with refined images than without, the effect is less pronounced than is for synthetic data. From this, we hypothesize that using synthetic data has a regularizing influence on the training with real data, namely that the learned features are more domain-invariant. This effect is smaller for images refined towards more realism.

\subsection{Analysis of Refinement Operations}
\begin{figure}
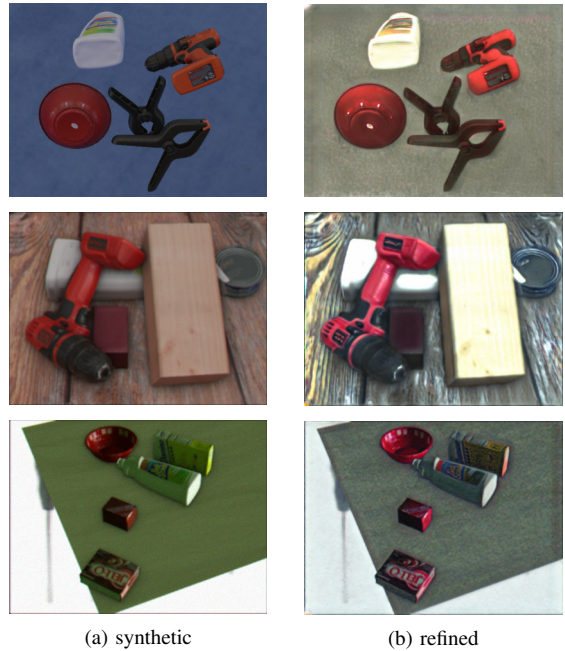

  \centering
    \centering\setlength\boxw{3.8cm}
    \begin{tikzpicture}[
       font=\footnotesize,
       b/.style={minimum width=\boxw, minimum height=3cm,inner sep=2pt},
       every outer matrix/.append={b}
     ]
     \node[b,matrix,matrix of nodes,nodes={draw=none,minimum width=0, minimum height=0},row sep=2pt, column sep=2pt] {
       \includegraphics[width=.9\boxw]{figures/cut_syn.png}   \\
       \includegraphics[width=.9\boxw]{figures/cut_syn_2.png} \\
       \includegraphics[width=.9\boxw]{figures/cut_syn_3.png} \\
       (a) synthetic \\
     };
    \end{tikzpicture}%
  \centering\setlength\boxw{3.8cm}
    \begin{tikzpicture}[
      font=\footnotesize,
      b/.style={minimum width=\boxw, minimum height=3cm,inner sep=2pt},
      every outer matrix/.append={b}
    ]
    \node[b,matrix,matrix of nodes,nodes={draw=none,minimum width=0, minimum height=0},row sep=2pt, column sep=2pt] {
      \includegraphics[width=.9\boxw]{figures/cut_refined.png}   \\
      \includegraphics[width=.9\boxw]{figures/cut_refined_2.png} \\
      \includegraphics[width=.9\boxw]{figures/cut_refined_3.png} \\
      (b) refined \\
    };
  \end{tikzpicture}%
  \vspace{-1ex}
  \caption{Synthetic images and their CUT-refined versions.}
  \label{fig:syn2real_example_syn}\label{fig:syn2real_example_refined}%
\end{figure}

We are mainly interested in achieving good segmentation performance to improve the value of synthetic training data for robotic applications. Still, it is worth analyzing what CUT is actually doing to the synthetic images. Figs.~\ref{fig:syn2real_example_syn}a) and b) show three synthetic images based on the YCB-Video objects and their CUT-refined versions, respectively.

While some refined images look more natural, some do not seem realistic to the human eye at all. Still, we achieve generalization to real data on the YCB-Video segmentation task that is close to what we get when training on real data. Hence, we hypothesize that---even if not for the human eye---refining the images aligns the synthetic and real image distributions more closely in the feature space of a CNN.

\begin{figure}
  \centering
  \includegraphics[width=\linewidth]{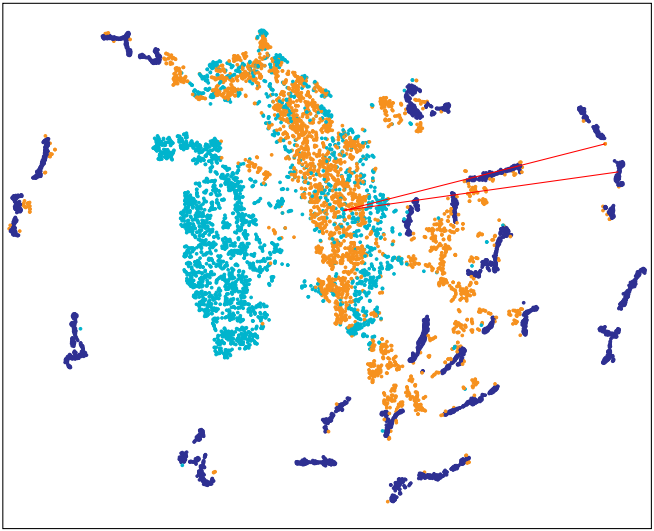}
  \caption{t-SNE embeddings for an early extraction layer of RefineNet (turquoise: synthetic, orange: refined synthetic, blue: real images). The red line connects a corresponding synthetic-refined-real tuple.}
  \label{fig:tsne_refinenet_1_2_conv1}
\end{figure}

\begin{figure}
  \centering
  \includegraphics[width=\linewidth]{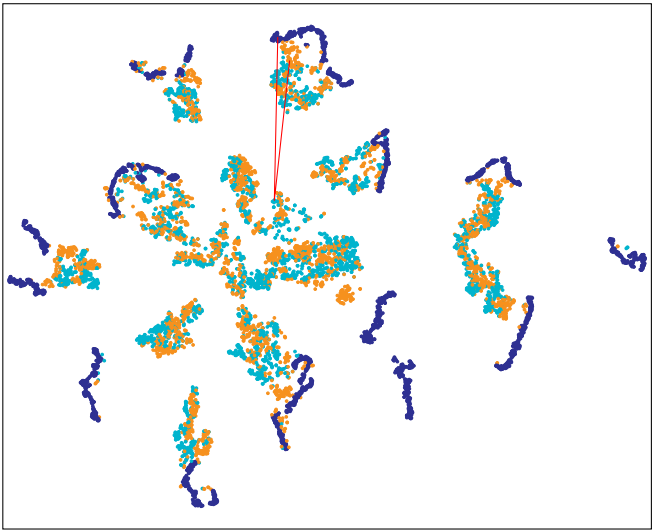}
  \caption{t-SNE embeddings for a late extraction layer of RefineNet (turquoise: synthetic, orange: refined synthetic, blue: real images). The red line connects a corresponding synthetic-refined-real tuple.}
  \label{fig:tsne_refinenet_3_2_conv1}
\end{figure}

As we cannot investigate this in the high-dimensional feature space directly, we employ t-SNE embeddings~\citep{van2008visualizing} to project the data into two dimensions. Specifically, we use the YCB-Video keyframes, render corresponding Stillleben images using the pose annotations and refine them using a trained CUT generator. As we are more interested in what happens to the appearance of the objects in the images rather than in the backgrounds, we mask out the backgrounds using the segmentation masks provided by Stillleben. For the resulting set of triples, we calculate the feature maps of one extraction layer of RefineNet trained on real YCB-Video images. We apply adaptive average pooling with output size 1$\times$1 to each feature map to obtain a vector of scalar values. Based on these vectors for all images, we calculate the t-SNE embeddings. The results vary based on the regarded layer. Embeddings based on features from an early extraction layer are depicted in \cref{fig:tsne_refinenet_1_2_conv1}. It can be seen that subsequent keyframes of the real YCB-Video video sequences are closely aligned. Besides, the synthetic images appear to form two clusters for which it was not possible to reliably determine their origin. We hypothesize that this split might be an artifact introduced by the tendency of \mbox{t-SNE} embeddings to form clusters, see~\citep{wattenberg2016tsne}. The general impression is that CUT-refining the images both spreads the distribution and also aligns the distribution closer to the real image distribution. For some images, the refined synthetic images are embedded quite close to the corresponding real images, as indicated by the exemplary red line in \cref{fig:tsne_refinenet_1_2_conv1}. In later extraction layers of RefineNet, the effect is less clearly visible but still present, see \cref{fig:tsne_refinenet_3_2_conv1}. We saw similar effects for the early layers of AlexNet~\citep{krizhevsky2012imagenet} trained on ImageNet\footnote{https://www.image-net.org/index.php}, supporting the hypothesis that the domain adaptation actually does align the distributions more closely.

\section{Discussion \& Conclusion}
We have presented a combined approach to generate training data from 3D object meshes using the synthesis pipeline Stillleben and unsupervised domain adaptation. We demonstrated the beneficial effect of our approach compared to purely synthetic data for a segmentation task on two robotics datasets, with only minor differences in the hyperparameters. For YCB-Video, the achieved segmentation performance is close to the performance with real training data. We explicitly remark that to apply our approach in new situations, like robotic competitions, only object meshes and images from real cameras are required, thus no annotation is necessary. With state-of-the-art GPU hardware, obtaining the necessary refined training data for a new environment is possible in far less than a day. From that, we conclude that our approach has the potential to be applicable in real robotic setups without requiring any real annotations.

A limitation inherited from Stillleben is the dependence on high-quality object meshes. Additionally, the performance we achieve is close to what is possible on real data but does not yet match it and the variation of the performance over the segmentation training is notably higher than for real images. Another point worth mentioning is that the analysis of the \mbox{t-SNE} embeddings only partly explains the good results; more research into this direction would be beneficial, also to find further room for improvement.

\section*{Acknowledgement}

This research has been funded by the Federal Ministry of Education and Research of Germany as part of the competence center for machine learning ML2R (01IS18038C).

\printbibliography

\end{document}